\pdfoutput=1
\documentclass[11pt]{article}

\usepackage[]{EMNLP2022}

\usepackage{times}
\usepackage{latexsym}
\usepackage{pifont}
\usepackage{bm}
\usepackage[T1]{fontenc}
\usepackage[utf8]{inputenc}
\usepackage{microtype}
\usepackage{amsthm,amsmath,amssymb}
\usepackage{mathrsfs}
\usepackage{booktabs}
\usepackage{multirow}
\usepackage{mathrsfs}
\usepackage{colortbl}
\usepackage{graphicx}  %插入图片的宏包
\usepackage{float}  %设置图片浮动位置的宏包
\usepackage{subfigure} 
\usepackage{soul}
\usepackage{dcolumn} 
\usepackage{hyperref}
\usepackage{siunitx}
\newcolumntype{d}[1]{D{.}{.}{#1}}
\definecolor{mygray}{gray}{.95}

\usepackage{inconsolata}

\title{Cross-domain Generalization for AMR Parsing}

\author{
 Xuefeng Bai\thanks{~~Work done as an intern at Tencent AI Lab Seattle.}$^{*\spadesuit}$\hspace{0.5mm},
 Sen Yang$^{\heartsuit}$\hspace{0.5mm}, 
 Leyang Cui$^{\clubsuit}$\hspace{0.5mm}, 
 Linfeng Song$^{\diamondsuit}$\hspace{0.5mm}, 
 Yue Zhang$^{\spadesuit \dagger}$\hspace{0.2mm}\hspace{1.5mm} \\
 $^\spadesuit$ School of Engineering, Westlake University, China\\
$^\heartsuit$ The Chinese University of Hong Kong, China \\
$^\clubsuit$ Tencent AI Lab, Shenzhen, China \\
 $^\diamondsuit$ Tencent AI Lab, Bellevue, WA, USA \\
 $^\dagger$ Institute of Advanced Technology, Westlake Institute for Advanced Study, China 
}

\begin{document}
\maketitle
\begin{abstract}
% This document is a supplement to the general instructions for *ACL authors. It contains instructions for using the \LaTeX{} style file for EMNLP 2022. 
% The document itself conforms to its own specifications, and is, therefore, an example of what your manuscript should look like.
% These instructions should be used both for papers submitted for review and for final versions of accepted papers.
% The document itself conforms to its own specifications, and is, therefore, an example of what your manuscript should look like.
% These instructions should be used both for papers submitted for review and for final versions of accepted papers.
% therefore, an example of what your manuscript should look like.
% These instructions should be used both for papers submitted for review and for final versions of accepted papers.
Abstract Meaning Representation (AMR) parsing aims to predict an AMR graph from textual input. 
Recently, there has been notable growth in AMR parsing performance.
However, most existing work focuses on improving the performance in the specific domain, ignoring the potential domain dependence of AMR parsing systems. 
To address this, 
we extensively evaluate five representative AMR parsers on five domains and analyze challenges to cross-domain AMR parsing.
We observe that challenges to cross-domain AMR parsing mainly arise from the distribution shift of words and AMR concepts.
Based on our observation, we investigate two approaches to reduce the domain distribution divergence of text and AMR features, respectively. 
Experimental results on two out-of-domain test sets show the superiority of our method.
\end{abstract}

\section{Introduction}

\begin{figure}
    \centering
    \includegraphics[width=0.35\textwidth]{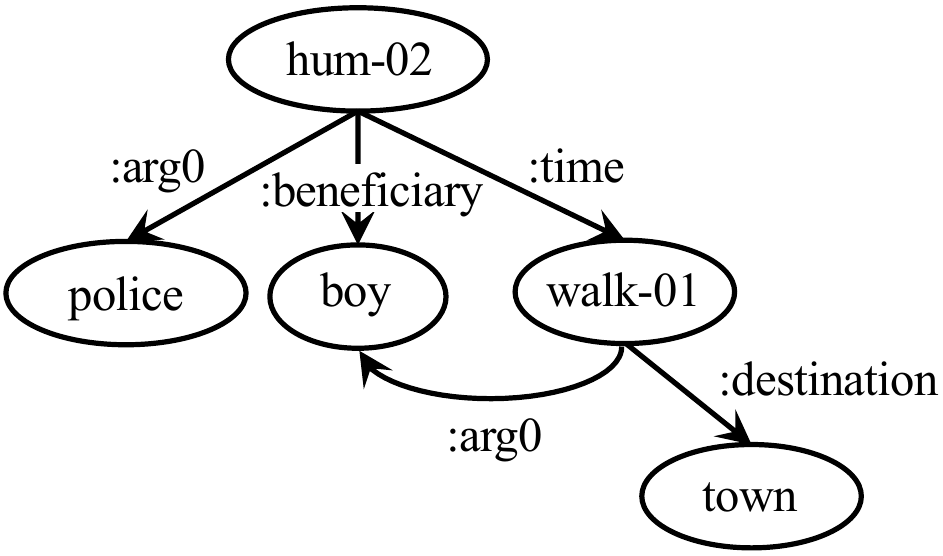}
    \caption{An AMR graph for sentence ``\textit{The police hummed to the boy as he walked to town.}''}
    \label{fig:intro-example}
    % \vspace{-1.0em}
\end{figure}

Abstract meaning representation (AMR; \citealt{banarescu2013abstract}) is a broad-coverage semantic structure formalism that represents the meaning of a text in a rooted directed graph.
As shown in Figure~\ref{fig:intro-example}, the nodes in an AMR graph represent concepts such as entities and predicates, and the edges indicate their semantic relations.
AMR parsing~\cite{flanigan-etal-2014-discriminative,konstas2017neural,TitovL18,guo-lu-2018-better,zhang-etal-2019-amr,cai-lam-2020-amr,Bevilacqua_Blloshmi_Navigli_2021,zhou-etal-2021-structure,bai-etal-2022-graph} is the task of transforming natural language into AMR graphs.
This is a fundamental task in semantics, which can also benefit downstream use.

AMR has been proven to be useful for many downstream tasks, such as information extraction~\cite{huang-etal-2016-liberal,Martinez-Rodriguez20,zhang-ji-2021-abstract,luo2022challenges,chen2022adaprompt,Wang-2022-WebFormer}, text summarization~\cite{liu-etal-2015-toward,liao2018abstract,chen-etal-2021-dialogsum,chen2022cross,he2022z},  machine translation~\cite{song2019semantic,slobodkin-etal-2022-semantics,Chen-2022-data}, text generation~\cite{konstas2017neural,song2018graph,zhu2019modeling,bai-etal-2020-online,ribeiro-etal-2021-investigating}, and dialogue systems~\cite{bonial-etal-2020-dialogue,bai-etal-2021-semantic,bai-etal-2022-semantic}.
To benefit such a diverse set of tasks that covers various domains, an ideal AMR parser should generalize well across different domains. 
However, most existing work only focuses on improving the in-domain parsing accuracy, ignoring the performances on other domains.
%, where the training and evaluation data are from the same domain.
Though state-of-the-art AMR parsers can obtain a \textsc{Smatch} score of over $84\%$ on an in-domain test set, we observe that their cross-domain performance is still weak (e.g., lower than $65\%$ on the biomedical domain).
It remains an open-question how well different types of AMR parsers generalize to out-of-domain (OOD) data.
% , and what the key factors are for AMR cross-domain generalization.
% }

In this work, we take the first step to study the cross-domain generalization ability of a range of typical AMR parsers, investigating three main research questions:
1) \textit{how well do different AMR parsers perform on out-of-domain test sets?} 2) \textit{what are the main challenges to cross-domain AMR parsing?} and 3) \textit{how to improve the performance of cross-domain AMR parsing?}

We empirically choose five major AMR parsers for comparison, including a two-stage statistical parser~\cite{flanigan-etal-2014-discriminative}, a graph-based parser~\cite{cai2020graph}, a transition-based parser~\cite{zhou-etal-2021-structure}, a Seq2Seq-based parser~\cite{Bevilacqua_Blloshmi_Navigli_2021}, and {an AMR-specific pre-training} parser~\cite{bai-etal-2022-graph}.
The test domains cover news, biomedical, novel, and wiki questions.
We conduct experiments under the zero-shot setting, where a model is trained on the source domain and evaluated on the target domain without using any target-domain labeled data.
Our results show that
1) all models give relatively lower (up to 45.5\%) performances on out-of-domain test sets, with the most dramatic drop on named entities and wiki links;
2) the graph pretraining-based parser is stronger in domain transfer than the other parsers;
% parsers that rely on complicated pre-processing and post-processing steps generalized bad in new domains (todo);
3) the transition-based parser is more robust than the seq2seq-based parser.
% In addition, fine-grained evaluation results show that the  are two main challenges for cross-domain AMR parsing. 
We further analyze the impact of a set of linguistic features, and the results suggest that the performance degradation is positively correlated with the distribution shifts of words and AMR concepts. 
Compared with the distribution divergences of the input features, those of the output features are more challenging to cross-domain AMR parsing.

Based on our analysis, 
%{we investigate two approaches for improving cross-domain AMR parsing.} 
we investigate two approaches to bridge the domain gap for improving cross-domain AMR parsing.
We first continually pre-train a BART model on target domain raw text to reduce the distribution gap of words. To further bridge the domain gap of output features, we adopt a pre-trained AMR parser to construct silver AMR graphs on the target domain, which potentially reduces the output features divergence. Experimental results show that the proposed methods consistently improve the parsing performance on out-of-domain test sets.
To our knowledge, this is the first systematic study on cross-domain AMR parsing.
Our code and results will be available at \url{https://github.com/goodbai-nlp/AMR-DomainAdaptation}.
% Based on our analysis, we propose to reduce distribution gap of words by continual pre-training a BART model on target domain raw text, and construct silver data to reduce the distribution divergence of AMR concepts, respectively.
% Experimental results show that the proposed method consistently improves the parsing performance on out-of-domain test sets.

% To our knowledge, this is the first systematically study on cross-domain AMR parsing. 

\section{Related Work}
\subsection{AMR Parsing}
% \bai{0622-night}
On a coarse-grained level, the current AMR parsing systems can be categorized into two main classes. 
The first is two-stage parsing system, which first identifies concepts, and then predicts relations based on the concept decisions. 
Two tasks are modeled either in a pipeline ~\cite{flanigan-etal-2014-discriminative, flanigan-etal-2016-cmu} or jointly ~\cite{TitovL18,zhang-etal-2019-amr}.
% For example, \citet{flanigan-etal-2014-discriminative, flanigan-etal-2016-cmu} rely on rule-based alignments and use a pipeline of concept and relation identification with a graph-based algorithm. 
% \citet{TitovL18} treat alignments as latent variables and solve alignments, concept and relation identification in a joint probabilistic model.
% \citet{zhang-etal-2019-amr} propose to jointly predict concepts and relations with an attention-based neural transducer.
The other one is one-stage parsing, which generates a
parse graph incrementally.
The one-stage parsing methods can be further divided into three categories:
graph-based parsing, transition-based parsing, and seq2seq-based parsing.
Transition-based parsing induces an AMR graph by predicting a sequence of transition actions.
The transition-based AMR parsers either maintain a stack and a buffer~\cite{wang-etal-2015-transition,damonte-etal-2017-incremental,ballesteros-al-onaizan-2017-amr,vilares-gomez-rodriguez-2018-transition,liu-etal-2018-amr,naseem-etal-2019-rewarding,fernandez-astudillo-etal-2020-transition,lee-etal-2020-pushing}
or make use of a pointer~\cite{zhou-etal-2021-amr,zhou-etal-2021-structure}.
Graph-based parsing builds a semantic graph incrementally. At each time step, a new node along with its connections to existing nodes are jointly decided.
The graph is induced either in top-down manner~\cite{cai-lam-2019-core} or in specific traversal order~\cite{zhang-etal-2019-broad,cai-lam-2020-amr}.
% For example,~\citet{cai-lam-2019-core} propose to a top-down graph induction strategy which first generates core concepts then digs more details. ~\citet{zhang-etal-2019-broad} transform the input text into a sequence of semantic relations, each of which includes a target node, a relation type, and a source node.
% ~\citet{cai-lam-2020-amr} iteratively complete the graph and update the representation of input sequence. 
Seq2seq-based parsing treats AMR parsing as a sequence-to-sequence problem by linearizing AMR graphs so that existing seq2seq models can be readily utilized.
Various seq2seq architectures have been employed for AMR parsing, such as vanilla seq2seq~\cite{barzdins-gosko-2016-riga,konstas2017neural}, supervised attention~\cite{peng-etal-2017-addressing}, character-based~\cite{van2017neural}, and pre-trained Transformer~\cite{Bevilacqua_Blloshmi_Navigli_2021,bai-etal-2022-graph}.

\begin{figure*}[t!]
	\centering 
	\subfigure[Two-stage parser]{\includegraphics[width=0.48\hsize]{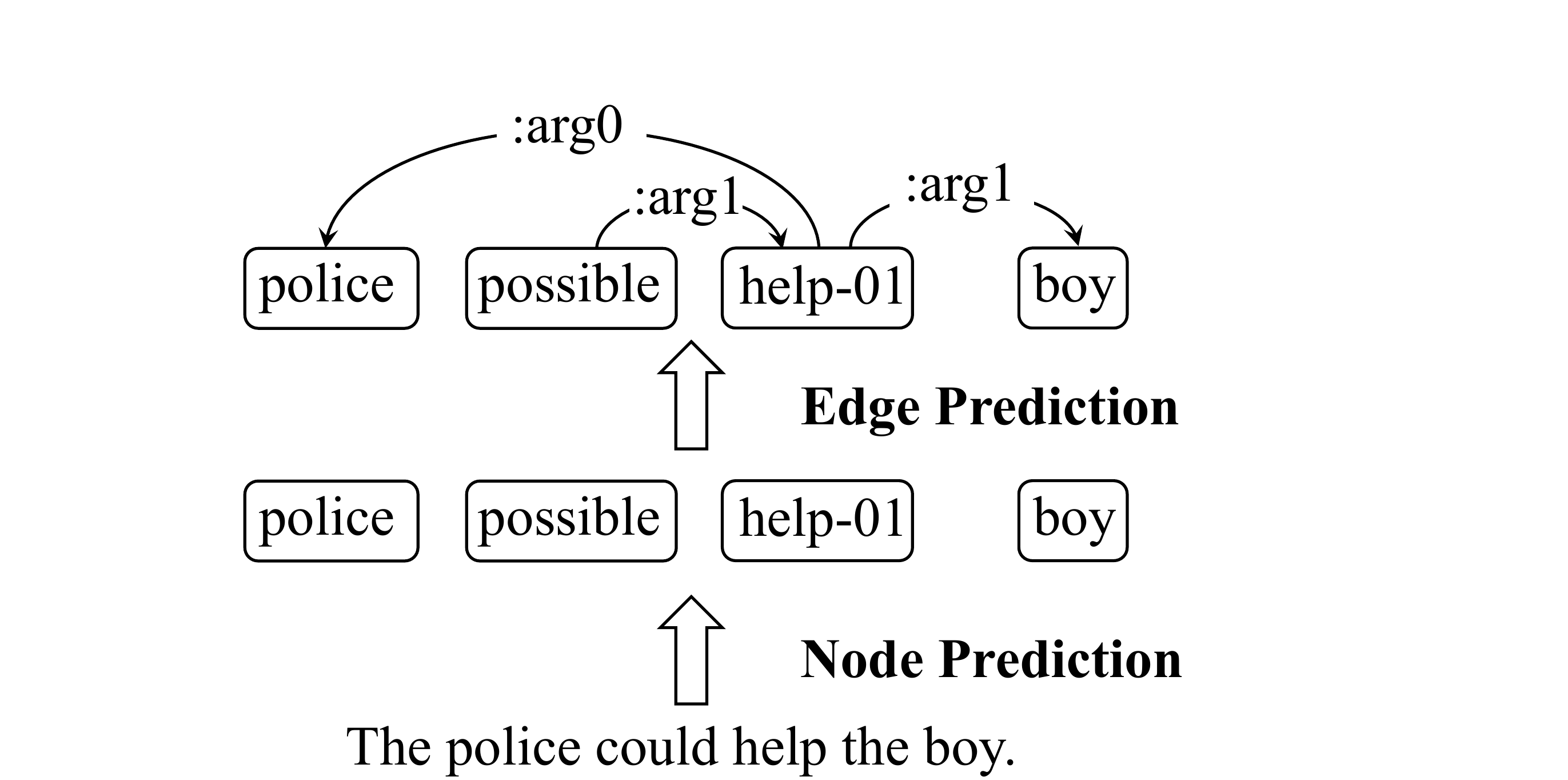}\label{fig:model1}} 
	\subfigure[Graph-based parser]{\includegraphics[width=0.48\hsize]{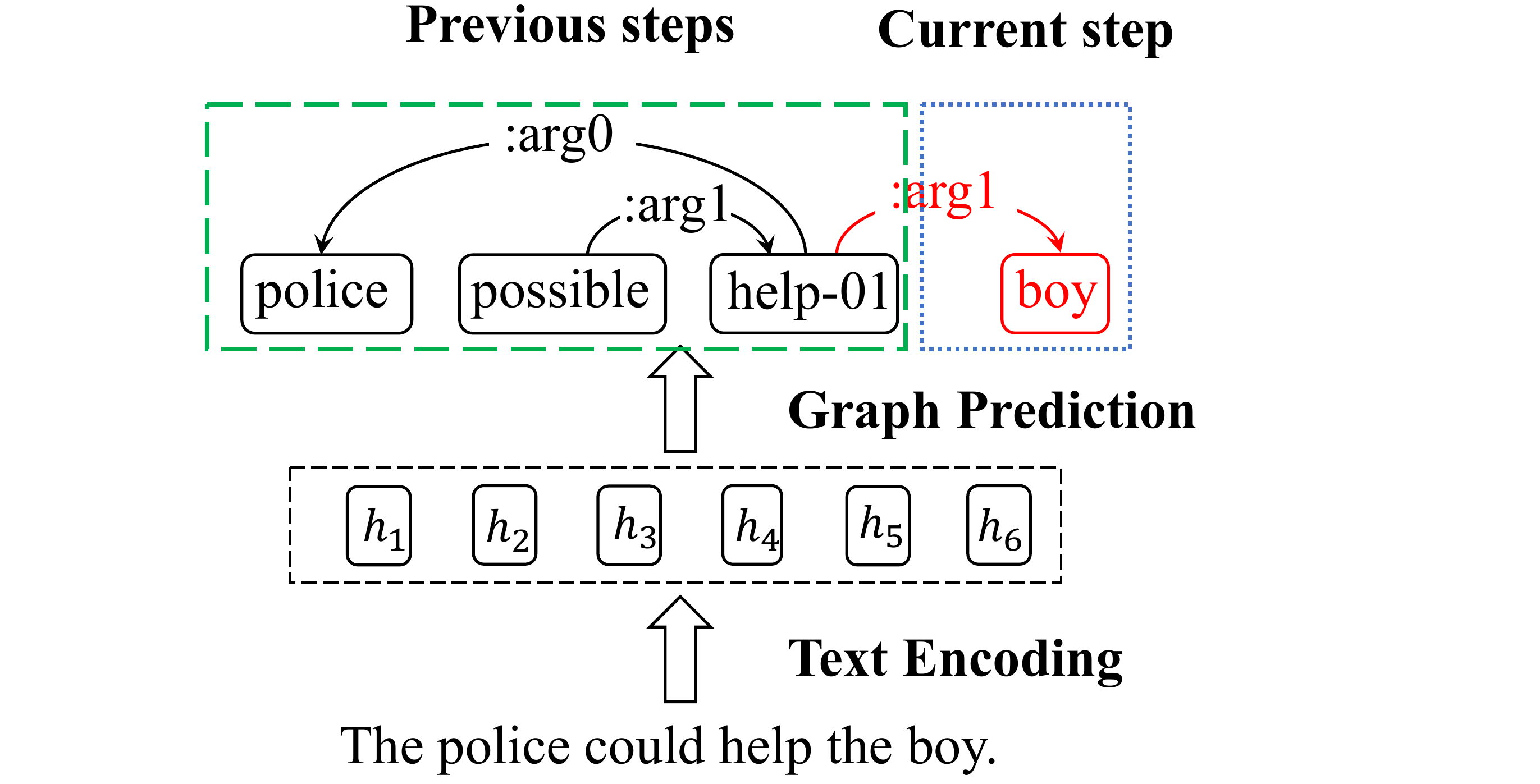}\label{fig:model2}}
	\subfigure[Transition-based parser]{\includegraphics[width=0.48\hsize]{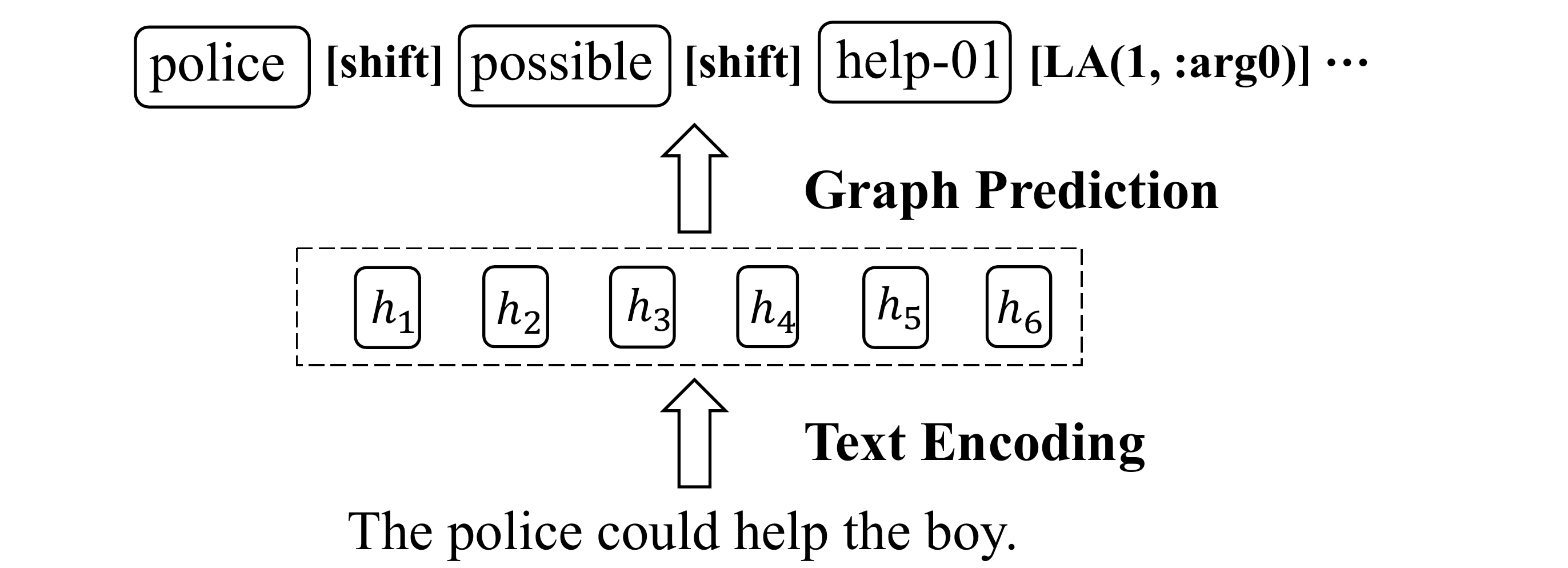}\label{fig:model3}}
	\subfigure[Seq2seq-based parser]{\includegraphics[width=0.48\hsize]{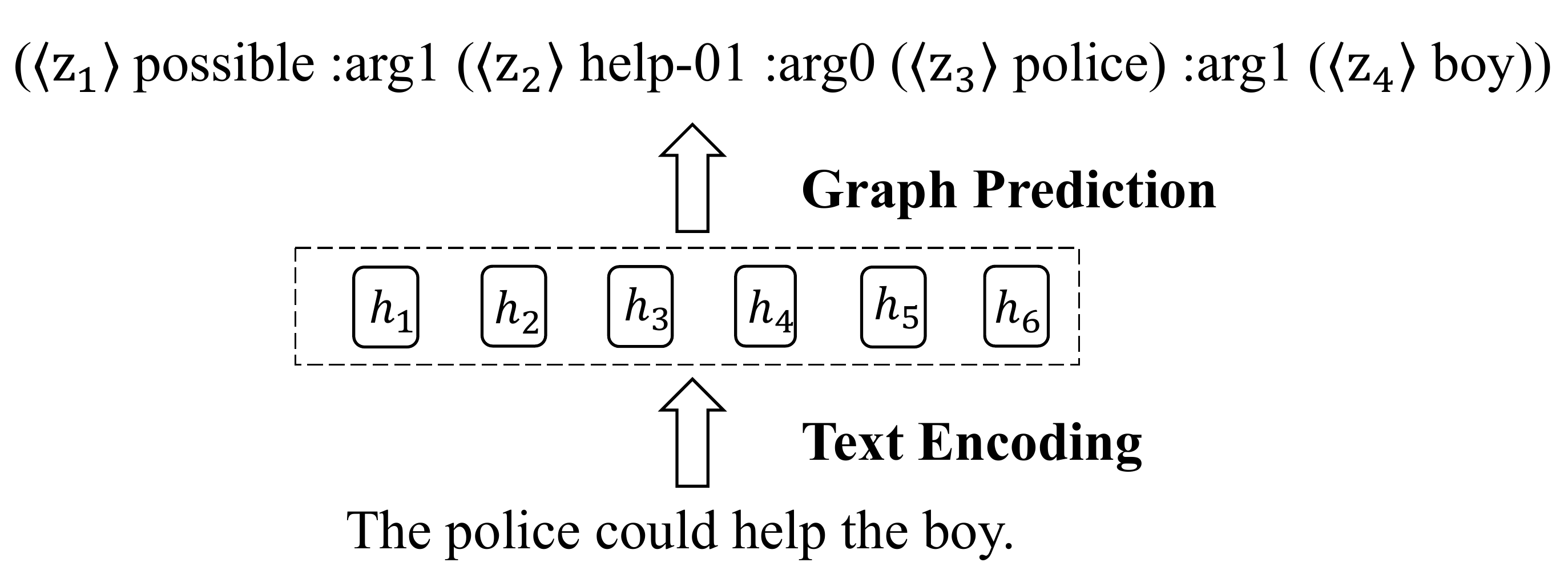}\label{fig:model4}}\\
	\caption{Illustration of four AMR parsers given the input ``\textit{The police could help the boy.}''.}
	\label{fig:model}
\end{figure*}

Despite great success, most previous work on AMR parsing focuses on the in-domain setting, where the training and test data share the same domain. 
In contrast, we systematically evaluate the model performance on $4$ out-of-domain datasets. 
To our knowledge, we are the first to systematically study cross-domain generalization for AMR parsing.

\subsection{Related Tasks}
% Named entity recognition (NER) and semantic role labeling (SRL) can be seen as semantic-related subtasks of AMR parsing.
% We thus summarize recent research studying the cross-domain generalization of these two tasks. 
We summarize recent research studying other semantic formalisms as well as the cross-domain generalization of named entity recognition (NER), semantic role labeling (SRL) and constituency parsing.

\noindent\textbf{Semantic parsing on other formalisms}. AMR is strong-correlated with other semantic formalisms such as semantic dependency parsing (SDP, \citealp{oepen-etal-2016-towards}) and universal conceptual cognitive annotation (UCCA, \citealp{abend-rappoport-2013-universal,hershcovich-etal-2017-transition}), and recent researches show that they can be represented in a unified format and parsed by a generalized framework~\cite{hershcovich-etal-2018-multitask,zhang-etal-2019-broad}. 
However, most of previous work focus on specific domain, leaving the study of cross-domain generalization unexplored.

\noindent\textbf{Cross-domain NER}. Named entity recognition (NER) is a subtask of AMR parsing.
To build a robust NER system across domains,
\citet{jointNER} directly train NER models on the domain-mixed corpus. 
\citet{wang-etal-2020-multi-domain-named} introduce an auxiliary task to predict the domain label. 
Recently, many studies focus on recognizing the unseen entity types in the target domain.
\citet{label-agnostic} and \citet{nearest-neighbor-crf} propose distance-based methods, which copy the entity label of nearest neighbors. 
% \citet{huang-etal-2021-shot} further pretrain distance-based model on noisy data. 
\citet{cui-etal-2021-template} and \citet{bert-ner} adopt prompt-based methods by using BART and BERT, respectively.

\noindent\textbf{Cross-domain SRL}. SRL can also be seen as semantic-related subtasks of AMR parsing.
\citet{dahlmeier2010domain} conduct an extensive study by analyzing various features and techniques that are used for SRL domain adaptation.
\citet{lim2014domain} combine a prior model with a structural learning model to build a multi-domain SRL system.
\citet{do-etal-15} exploit the knowledge from a neural language model and external linguistic resource for domain adaptation on biomedical data.
% \citet{hartmann-etal-2017-domain} create an out-of-domain test set for FrameNet SRL and use distributed word representations to improve out-of-domain SRL performance.
\citet{rajagopal-etal-2019-domain} develop a label mapping strategy and a layer adapting approach for cross-domain SRL. 
Compared with cross-domain NER and SRL, the task of cross-domain AMR parsing is more challenging since AMR is a graph formalism, and AMR contains more types of concepts and relations.

\noindent\textbf{Cross-domain constituency parsing.} ~\citet{yang-etal-2022-challenges} investigated challenges to open-domain syntactic parsing, introducing datasets on new domains and analyzing the key factors on to cross-domain constituency parsing using a set of linguistic features. 
Our work is similar to their work in studying the key challenges on various parsing systems.
However, we focus on AMR and conducts fine-grained semantic-related evaluation. 
In addition, we provide a intuitive solution for improving cross-domain AMR parsing.

\section{Compared Models}
% Our choice of parsers for evaluation is motivated by the broad family  
% Generally, current AMR parsing methods can be categorized into two families: Two-stage parsing and One-stage parsing. 
% The latter can be further divided into three classes: graph-based, transition-based and sequence-to-sequence based. 
We choose the representative or top-performing parser of two-stage, graph-based, transition-based, seq2seq-based as well as a pre-trained parser for evaluation.
In particular, the following AMR parsing systems are considered:

\begin{table*}[!t]
	\centering
	\small
	\begin{tabular}{lccccc}
		\toprule
		\textbf{Models} & \textbf{Categratory} & \textbf{Pre-proc.} &\textbf{Post-proc.} &\textbf{Ext. Data} &\textbf{PLM}  \\
		\midrule
		JAMR &Two-stage & $\checkmark$ &$\checkmark$ &POS, train align, etc. &\ding{55} \\
		AMRGS &Graph &Recat. &concept, polarity, wiki & POS, NER, Lemm. &\textsc{BERT} \\
		\textsc{StructBART} &Transition &\ding{55} &wiki &train align. &BART \\
		SPRING &Seq2seq  &\ding{55} &wiki &\ding{55} &BART \\
		AMRBART &Pretrain + Seq2seq &\ding{55} &wiki &200k silver &BART \\
		\bottomrule
	\end{tabular}
	\caption{Compared AMR parsing systems. ``Recat''--graph re-categorization.}
	\label{tab:models}
% 	\vspace{-0.5em}
\end{table*}

\textbf{JAMR}~\cite{flanigan-etal-2014-discriminative}, as shown in Figure~\ref{fig:model1}, is a two-stage parsing model which predicts concepts and relations in a pipeline.
JAMR identifies concepts and predicts the relations using two discriminatively-trained linear structured predictors, which use rich features like part-of-speech tagging (POS), named entities recognition (NER), lemmatization, etc.
In addition, JAMR relies on an external aligner to construct supervision signals for both stages.

\textbf{AMRGS}~\cite{cai-lam-2020-amr} is a graph-based parser which builds a semantic graph incrementally.
As shown in Figure~\ref{fig:model2}, at every step, the graph-based parser predicts one node and its connection to existing graph.
AMRGS learns mutual causalities between text and graph by updating the sentence and graph representations iteratively.
AMRGS obtains word-level representation from a pre-trained language model (i.e., BERT~\cite{devlin-etal-2019-bert}) and  uses POS, NER and lemmatization as external knowledge to make predictions.

\textbf{\textsc{StructBART}}~\cite{zhou-etal-2021-structure}, as shown in Figure~\ref{fig:model3}, is a transition-based parser which generates an AMR graph through a sequence of transition actions. 
In particular, the transition actions are:

\indent\textbullet~\textbf{\textsc{shift}} moves token cursor to right. \\
\indent\textbullet~\textbf{<string>} creates a node of name <string>. \\
\indent\textbullet~\textbf{\textsc{copy}} creates a node with the name of the \indent cursor-pointed token. \\
\indent\textbullet~\textbf{\textsc{la}(j, \textsc{lbl})} creates an \textit{arc} with label \textsc{lbl} from 
\indent the last generated node to the jth generated node. \\
\indent\textbullet~\textbf{\textsc{ra}(j, \textsc{lbl})} is same as \textsc{la} but with reversed \indent edge direction. \\
\indent\textbullet~\textbf{\textsc{root}} assigns the last generated node as root.

StructBART takes a pre-trained BART model as the backbone and extends the original vocabulary with transition actions.
Additionally, StructBART requires an external aligner to obtain oracle transition actions for training.

\textbf{\textsc{SPRING}}~\cite{Bevilacqua_Blloshmi_Navigli_2021}, as shown in Figure~\ref{fig:model4}, is a sequence-to-sequence parser which transforms a text sequence into a linearized AMR sequence.
{SPRING} adopts a depth-first algorithm to transform AMR graphs into a sequence where concepts and relations are treated equally.
To deal with co-referring nodes, SPRING adds special tokens to the vocabulary.
Same with \textsc{StructBART}, SPRING also initializes model parameters with BART.

\textbf{AMRBART}~\cite{bai-etal-2022-graph} is
% a sequence-to-sequence parser augmented with pre-trained structure-aware encoder and decoder. 
% AMRBART continually pre-trains a BART model on text, AMR graph and joint AMR and text using $6$ pre-training tasks.
a continually pre-trained BART model on AMR graphs and text.
It uses three graph-based pre-training tasks to improve the structure awareness of the encoder and decoder and another four tasks that jointly learns on text and AMR graph to capture the correspondence between AMR and text.
AMRBART is pre-trained on 250k training instances, which lie in the same domain as AMR2.0.

In addition, JAMR uses complicated rule-based pre-processing and post-processing steps to simplify the input and reconstruct the AMR graphs.
AMRGS uses rule-based graph re-categorization for pre-processing and recovers concept sense tags, wiki links, and polarities during post-processing.
StructBART, SPRING, and AMRBART do not require pre-processing steps and use the BLINK Entity Linker~\cite{wu-etal-2020-scalable} to handle wiki links during post-processing.
Table~\ref{tab:models} summarizes the above systems according to their characteristics. 

\section{Experiments}
Experimental configurations and our adopted datasets are shown in Sections~\ref{sec:experimental_setting} and \ref{sec:datasets}, respectively.
To study the cross-domain generalization ability of current AMR parsers, we first quantify the difference between in-domain training data and out-of-domain test data (Section~\ref{sec:Discrepancy}), and then evaluate the cross-domain performance of $5$ typical AMR parsers (Section~\ref{sec:cross-domain-res}).

\subsection{Experimental Settings}
\label{sec:experimental_setting}
\noindent\textbf{Model Configuration}. 
We adopt the officially released code of each system and use their default configuration to re-train and evaluate the model performance.
The best model is selected according to the performance on the in-domain validation set. 
All models are trained and evaluated on a single Nvidia Tesla V100 GPU.

\noindent\textbf{Metrics.} We assess the performance of parsing models with \textsc{Smatch}~\cite{cai-knight-2013-smatch} scores computed with the \textit{amrlib}\footnote{\url{https://github.com/bjascob/amrlib}} tools, which also report fine-grained scores including unlabeled, NoWSD, concept identification, NER, negations, reentrancy and wiki links.\footnote{Please refer to appendix~\ref{appendix:metric} for detailed definitions.}

\begin{table}
	\centering
	\small
	\begin{tabular}{l|c|c|c}
		\toprule
        Dataset & Category & Sents & Tokens \\
		\midrule
		\multicolumn{4}{l}{\textbf{ID AMRs}} \\
		\midrule
		AMR2.0 & train & 36,521 & 653K \\
		       & dev & 1,371 & 30K \\
		       & test & 1,368 & 29K \\
        \midrule
        \multicolumn{4}{l}{\textbf{OOD AMRs}} \\
        \midrule
        New3 & train & 4,441 & 83K \\
            & dev & 354 & 64K \\
		    & test & 527 & 8K \\
		\midrule
        TLP & test &1,562 & 21K \\
        \midrule
        Bio & train & 5,452 & 138K \\
		    & test & 500 & 13K \\
        \midrule
		QALD-9 & test & 558 & 5K \\
		\midrule
		\multicolumn{4}{l}{\textbf{External Data}} \\
		\midrule
		Raw text (TLP) & - &109k & 2M \\
		Raw text (Bio) & - &200k & 4.4M \\
		Silver AMR (TLP) & - &109k &2M \\
		Silver AMR (Bio) & - &200k &4.4M \\
	   \bottomrule
	\end{tabular}
	\caption{Dataset statistics.}
	\label{tab:datasets}
% 	\vspace{-1.8em}
\end{table}

\begin{table*}
	\centering
	\small
	\begin{tabular}{lccccccc}
		\toprule
		\multirow{2}{*}{Datasets} & \multicolumn{4}{c}{Input Features}  & \multicolumn{3}{c}{Output Features}\\
		\cmidrule(lr){2-5} \cmidrule(lr){6-8}
         & Avg. Len & Unigram & Bigram & Trigram & Concept & Relation & Triplet \\
		\midrule
		AMR2.0 (ID) &19.76 & 0.14 (0.05) &0.46 (0.39) & 0.64 (0.78) &0.14 (0.04) & 0.07 (0.00) &0.43 (0.32)  \\
        \midrule
        New3 &14.69 & 0.24 (0.10) & 0.57 (0.50) & 0.68 (0.85) &0.26 (0.10) &0.03 (0.00) & 0.53 (0.44) \\
        TLP &13.69 & 0.22 (0.04) & 0.51 (0.34) & 0.66 (0.78) & 0.30 (0.05) & 0.06 (\textbf{1e-3}) & 0.60 (0.57) \\
        Bio &\textbf{25.20} & \textbf{0.39} (\textbf{0.29}) & 0.63 (\textbf{0.78}) & 0.69 (\textbf{0.95}) & \textbf{0.44} (\textbf{0.21}) &0.06 (\textbf{1e-3}) & \textbf{0.66} (\textbf{0.71}) \\
        QALD-9 &~7.52 & 0.38 (0.08) & \textbf{0.64} (0.48) & \textbf{0.69} (0.84) &0.38 (0.08)  & \textbf{0.07} (0.00) & 0.58 (0.40) \\
	   \bottomrule
	\end{tabular}
	\caption{Feature difference between AMR2.0 training set and $5$ test sets. We report the Jensen-Shannon divergence (and OOV rate) for features except input length. A lower JS divergence (and OOV) value means that the test set is more similar to AMR2.0 training set on that specific feature.}
	\label{tab:dataset-diff}
% 	\vspace{-10pt}
\end{table*}

\subsection{Datasets}
\label{sec:datasets}
\noindent\textbf{In-Domain Dataset}. We train and evaluate AMR parsers on standard benchmarks, which we refer to as the In-Domain (ID) setting. We use \textbf{AMR2.0} (LDC2017T10)\footnote{\url{https://catalog.ldc.upenn.edu/LDC2017T10}} as ID dataset which consists AMRs from newswire, discussion forum and other web logs, web collections.

\noindent\textbf{Out-of-Domain Datasets}. We consider the following datasets for out-of-domain (OOD) evaluation: \textbf{New3}, a subset of AMR3.0\footnote{\url{https://catalog.ldc.upenn.edu/LDC2020T02}}, whose original source was the DARPA LORELEI program~\cite{LORELEI18}. 
The domain of New3 is close to AMR2.0; 
\textbf{TLP}\footnote{\url{https://amr.isi.edu/download/amr-bank-struct-v1.6.txt}} is an annotation of the novel~\textit{The Little Prince} that contains 1,562 sentences.
% We use the original version (\textit{ver}. 1.6) for evaluation.
\textbf{Bio}\footnote{\url{https://amr.isi.edu/download/2016-03-14/amr-release-test-bio.txt}}, which consists of annotations of biomedical texts, including PubMed articles and sentences from other biological corpus.
% We report results based on the updated version (\textit{ver.} 3.0)\footnote{\url{https://amr.isi.edu/download/2018-01-25/amr-release-bio-v3.0.txt}} of Bio.
% We follow the original version (\textit{ver.} 0.8) to split train/dev/test dataset, but use the latest corpus (\textit{ver.} 3.0)\footnote{\url{https://amr.isi.edu/download/2018-01-25/amr-release-bio-v3.0.txt}} for evaluation.
\textbf{QALD-9}\footnote{\url{https://github.com/IBM/AMR-annotations}}~\cite{youngsuklee-etal-2022-naacl2022}, a recent released dataset whose original source the questions of SQuAD2.0~\cite{SQuAD16}. 
Since~{QALD-9} comprises only 150 test sentences, we concatenate the train and test set for evaluation, leading to 558 instances in total.

In addition, we collect raw text from two domains: biomedical data (like Bio) and fairy tales data (like TLP). 
The former is sampled from  PubMedQA~\cite{jin-etal-2019-pubmedqa} dataset, while the latter is a collection of fairy tales between the 19th century and early 20th.
We also construct silver data for TLP and Bio by employing a state-of-the-art AMR parser~\cite{bai-etal-2022-graph} to parse collected raw sentences into AMR graphs.
Table~\ref{tab:datasets} shows more details of above datasets.

\subsection{Distributional Variance Across Datasets}
\label{sec:Discrepancy}

%%%%%%%%%%%%%%%%%  Results on TLP v3.0 and BIO v3.0 %%%%%%%%%%%%%%%%%%%%%%%%%
% \begin{table*}
% 	\centering
% 	\small
% 	\begin{tabular}{lcccccc}
% 		\toprule
% 		\multirow{2}{*}{\textbf{Model}} & \multicolumn{1}{c}{\textbf{ID}} & \multicolumn{5}{c}{\textbf{OOD}} \\
% 		\cmidrule(lr){2-2} \cmidrule(lr){3-7}
%          & \textbf{AMR2.0} & \textbf{New3}  & \textbf{TLP} & \textbf{Bio} &\textbf{QALD-9} & \textbf{Avg} \\
% 		\midrule
% 		JAMR &67.0 & 57.2 (14.6\%) &58.8 (12.2\%) &38.4 (42.7\%) &60.8 (9.3\%) & 53.8 (19.7\%) \\
% 		AMRGS & 80.6 & 61.8 (23.3\%) & 72.0 (10.7\%)	& 43.2 (46.4\%) & 70.0 (13.1\%) &61.8 (23.3\%) \\
% 		\textsc{StructBART} &84.1 & 74.0 (12.0\%) &78.5 (6.7\%) &57.6 (31.5\%) &83.7 (\textbf{0.5\%}) & 73.5 (12.6\%) \\
% 		SPRING &84.7  &74.2 (12.2\%) &78.2 (7.5\%) &59.2 (29.9\%) & 80.4 (4.9\%) & 73.0 (13.8\%)\\
% 		AMRBART &\textbf{85.5} &\textbf{77.3} (\textbf{9.6\%}) &\textbf{79.8} (\textbf{6.7\%}) &\textbf{62.1} (\textbf{27.4\%}) & \textbf{85.1} (\textbf{0.5\%}) &\textbf{76.1} (\textbf{11.0\%}) \\
% 		\bottomrule
% 	\end{tabular}
% 	\caption{\textsc{Smatch} scores on in-domain (ID) and out-of-domain (OOD) test sets and the relative performance reduction rate for OOD test sets. The best results within each column are shown in \textbf{bold}.}
% 	\label{tab:main-parsing}
% % 	\vspace{-0.7em}
% \end{table*}

\begin{table*}
	\centering
	\small
	\begin{tabular}{lcccccc}
		\toprule
		\multirow{2}{*}{\textbf{Model}} & \multicolumn{1}{c}{\textbf{ID}} & \multicolumn{5}{c}{\textbf{OOD}} \\
		\cmidrule(lr){2-2} \cmidrule(lr){3-7}
         & \textbf{AMR2.0} & \textbf{New3}  & \textbf{TLP} & \textbf{Bio} &\textbf{QALD-9} & \textbf{Avg} \\
		\midrule
		JAMR &67.0 & 57.2 (14.6\%) &59.9 (11.9\%) &38.7 (42.2\%) &60.8 (9.3\%) & 54.2 (19.2\%) \\
		AMRGS & 80.6 & 61.8 (23.3\%) & 73.7 ( 9.4\%)	& 43.9 (45.5\%) & 70.0 (13.1\%) &62.4 (22.6\%) \\
		\textsc{StructBART} &84.1 & 74.0 (12.0\%) &80.2 (4.9\%) &60.4 (28.2\%) &83.7 (\textbf{0.5\%}) & 74.6 (11.3\%) \\
		SPRING &84.7  &74.2 (12.2\%) &79.9 (6.0\%) &59.7 (29.5\%) & 80.4 (4.9\%) & 73.6 (13.2\%)\\
		AMRBART &\textbf{85.5} &\textbf{77.3} (\textbf{9.6\%}) &\textbf{81.6} (\textbf{4.8\%}) &\textbf{63.2} (\textbf{26.1\%}) & \textbf{85.1} (\textbf{0.5\%}) &\textbf{76.8} (\textbf{10.2\%}) \\
		\bottomrule
	\end{tabular}
	\caption{\textsc{Smatch} scores on in-domain (ID) and out-of-domain (OOD) test sets and the relative performance reduction rate for OOD test sets. The best results within each column are shown in \textbf{bold}.}
	\label{tab:main-parsing}
% 	\vspace{-10pt}
\end{table*}

To better understand the cross-domain parsing performance, we quantify the difference between the ID training set and $5$ test sets according to the following list of linguistic features: \textbf{input text features}, including input length, uni-gram, bi-gram, tri-gram;\textbf{ output AMR features}, which consists of AMR concepts, AMR relations, and $\langle \text{concept}, \text{relation}, \text{concept} \rangle$ triplets.
We report the average score for input length.
For other features, we follow~\citet{yang-etal-2022-challenges} to consider both the out-of-vocabulary (OOV) rate and the Jensen-Shannon divergence~\cite{Jensen-Shannon04} to measure the difference.
The former calculates the vocabulary difference between two domains, while the latter records the distributional divergence.
Given a specific feature, denoting the feature distribution in the source domain as $P$ and the distribution in the target domain as $Q$, the Jensen-Shannon divergence (\textit{JS}) is calculated as:
\begin{equation}
\begin{split}
    \textit{JS}(P||Q) &= \frac{1}{2}(\textit{KL}(P||M) + \textit{KL}(Q||M)), \\
    M &= \frac{1}{2}(P+Q), \\
\end{split}
\end{equation}
where \textit{KL} represents the Kullback-Leibler divergence~\cite{csiszar1975divergence}.
A lower \textit{JS} divergence value means that the test set is more similar to AMR2.0 training set on that specific feature.

As shown in Table~\ref{tab:dataset-diff}, the main difference between the test sets and the training set comes from the input length, the unigram/bigram/trigram, the concept, and the triplet, while the relation difference is relatively small.
The vocabulary differences (e.g., unigram OOV rate and concept OOV rate) are relatively small compared with feature distribution divergence.
% Also, the concept difference is positively related to unigram difference.
Among all the test sets, AMR2.0 is the closest to the training set.
In contrast, Bio has the longest average input length and the largest overall feature difference from the AMR2.0 training data. 
In particular, the unigram OOV rate of Bio is 0.29, which is much bigger than that of other test sets.
New3 and TLP have medium input length, and the feature differences are smaller than Bio.
QALD-9 has an average input length of $7.5$, which is $2.5$ times smaller than that of AMR2.0. 
Overall, we can observe that individual statistics vary across domains, which reflects large domain differences.

\subsection{Cross-Domain AMR Parsing Performance}
\label{sec:cross-domain-res}
Table~\ref{tab:main-parsing} lists the performances of $5$ parsers on $5$ domains.
All models achieve their best results on the in-domain AMR2.0 test set.
By contrast, the performance drops on OOD test sets, ranging from $0.5\%$ to $45.5\%$, showing that cross-domain AMR parsing is still a challenge. 

Among all domains, Bio is the hardest one, which is in line with our observation in feature differences (i.e., Table~\ref{tab:dataset-diff}). 
The \textsc{Smatch} scores of all parsers on Bio fall to a range between $38.7$ and $63.2$, which is much lower than those on ID test sets (from $67.0$ to $85.5$). 
The reason can be two-fold: 
First, Bio contains many biomedical terminologies, resulting in significant feature differences. For example, the unigram OOV rate is $29\%$, and the JS divergence of concept is $0.44$ on Bio, according to Table~\ref{tab:dataset-diff}.
Second, the average input text length of Bio is larger than those of other test sets.
In comparison, QALD-9 is the easiest, with a relative performance reduction rate ranging from $0.5\%$ to $13.1\%$.
The main reason could be that the input text length of QALD-9 is small, which significantly reduces the difficulty of AMR parsing.
We give further analysis on these features in Section~\ref{sec:analysis}.

Among all the systems, AMRBART gives the best \textsc{Smatch} scores on all test sets and has the lowest relative performance reduction rates, indicating that large-scale graph pre-training, which has been shown to boost ID performance, is also helpful for improving OOD generalization.
SPRING gives better results than \textsc{StructBART} on the ID test set and a bigger relative performance reduction rate on OOD test sets, which may result from the fact that the transition-based model implicitly learns the local correspondence between AMR and text, which is helpful for generalization. In contrast, the seq2seq-based model focus on sequence-level transduction.
Comparing AMRGS with other models, though all four neural parsers achieve \textsc{Smatch} scores of over $80.0$, AMRGS shows much lower OOD performances than the other three neural parsers, and some of its OOD relative performance reduction rates are even bigger than those of the non-neural JAMR parser. 
This might be caused by the difference on rule-based processing methods for AMR graphs: as shown in Table~\ref{tab:models}, AMRGS utilizes much more rule-based methods to pre- (and post-) process AMR graphs than the other three neural methods do. Since these rules are derived from the training data, such domain-specific rules would not generalize well to new domains. 
We also give the full evaluation results, please refer Appendix~\ref{sec:appendix-more-res}.

\begin{figure}
    \centering
    \includegraphics[width=1.0\hsize]{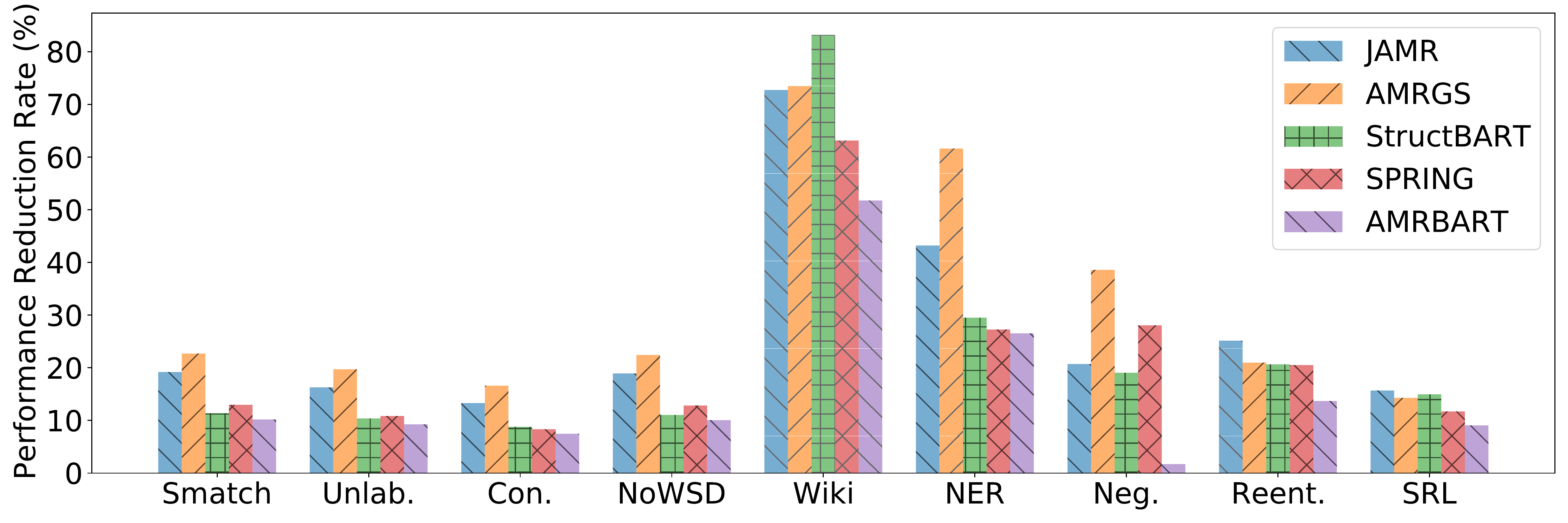}
    \caption{Relative performance reduction rate in terms of different evaluation metrics. ``Unlab.''--unlabeled, ``Con.''--Concept, ``Neg''--Negation, ``Reent.''--Reentrancy.}
    \vspace{-1em}
    \label{fig:metrics}
\end{figure}

\section{Key Challenges to OOD AMR Parsing}
\label{sec:analysis}
Based on the results above, we further study two important questions: \textit{which AMR components are the most challenging for cross-domain AMR parsing (in Section~\ref{sec:error-analysis})}; and \textit{what contributes most to the performance degradation on OOD test sets (in Section~\ref{sec:factors})}?

\subsection{Error Analysis} % 从模型表现层面，待修改
\label{sec:error-analysis}
Figure~\ref{fig:metrics} gives the relative performance reduction of each model regarding \textsc{Smatch} and $8$ fine-grained evaluation metrics (measured by F1 scores). 
We report the average score of four OOD test sets.
Among all evaluation metrics, wiki links and named entities are the hardest objectives to handle.
This is intuitive because a large proportion of wiki links and named entities contain out-of-vocabulary (OOV) tokens.
Also, there are OOV named entity types in the OOD test sets. 
For example, 3.7\% of named entity types of Bio are never seen during training, which further increases the difficulty level for AMR parsers to predict the correct labels.
The performance of negative polarity and reentrancy detection also drops significantly, with average scores of $22\%$ and $20\%$, respectively. 
For unlabeled\footnote{\textsc{Smatch} while ignoring relation labels.}, concept, NoWSD\footnote{\textsc{Smatch} while ignoring Propbank senses.} and semantic role labeling (SRL), we observe relatively lower performance degradation.
% \begin{table*}
% 	\centering
% 	\small
% 	\begin{tabular}{lcccccccc}
% 		\toprule
% 		\multirow{2}{*}{Datasets} & \multicolumn{5}{c}{Input Features}  & \multicolumn{3}{c}{Output Features}\\
% 		\cmidrule(lr){2-6} \cmidrule(lr){7-9}
%          & Avg. Len & 1-gram & 2-gram & 3-gram & 4-gram & Node & Edge & Triple \\
% 		\midrule
% 		AMR2.0 &19.76 & 0.14 (0.05) &0.46 (0.39) & 0.64 (0.78) & 0.68 (0.94) &0.14 (0.04) & 0.07 (0.00) &0.43 (0.32)  \\
%         \midrule
%         New3 &14.69 & 0.24 (0.10) & 0.57 (0.50) & 0.68 (0.85) & 0.69 (0.97) &0.26 (0.10) &0.03 (0.00) & 0.53 (0.44) \\
%         TLP &13.69 & 0.22 (0.04) & 0.51 (0.34) & 0.66 (0.78) & 0.69 (0.96) & 0.30 (0.05) & 0.06 (\textbf{1e-3}) & 0.60 (0.57) \\
%         Bio &\textbf{25.20} & \textbf{0.39} (\textbf{0.29}) & 0.63 (\textbf{0.78}) & 0.69 (\textbf{0.95}) & 0.69 (\textbf{0.99}) & \textbf{0.44} (\textbf{0.21}) &0.06 (1e-3) & \textbf{0.66} (\textbf{0.71}) \\
%         QALD-9 &~7.52 & 0.38 (0.08) & \textbf{0.64} (0.48) & \textbf{0.69} (0.84) & \textbf{0.69} (0.97) &0.38 (0.08)  & \textbf{0.07} (0.00) & 0.58 (0.40) \\
% 	   \bottomrule
% 	\end{tabular}
% 	\caption{todo.}
% 	\label{tab:dataset-diff}
% % 	\vspace{-1.8em}
% \end{table*}

% \begin{figure*}
%     \centering
%     \includegraphics[width=1.1\hsize]{all-metrics.pdf}
%     \caption{Performance decrease regarding to evaluation metrics}
%     \label{fig:metrics}
% \end{figure*}

\begin{figure}[!t]
    \centering
	\subfigure[]{\includegraphics[width=0.9\hsize]{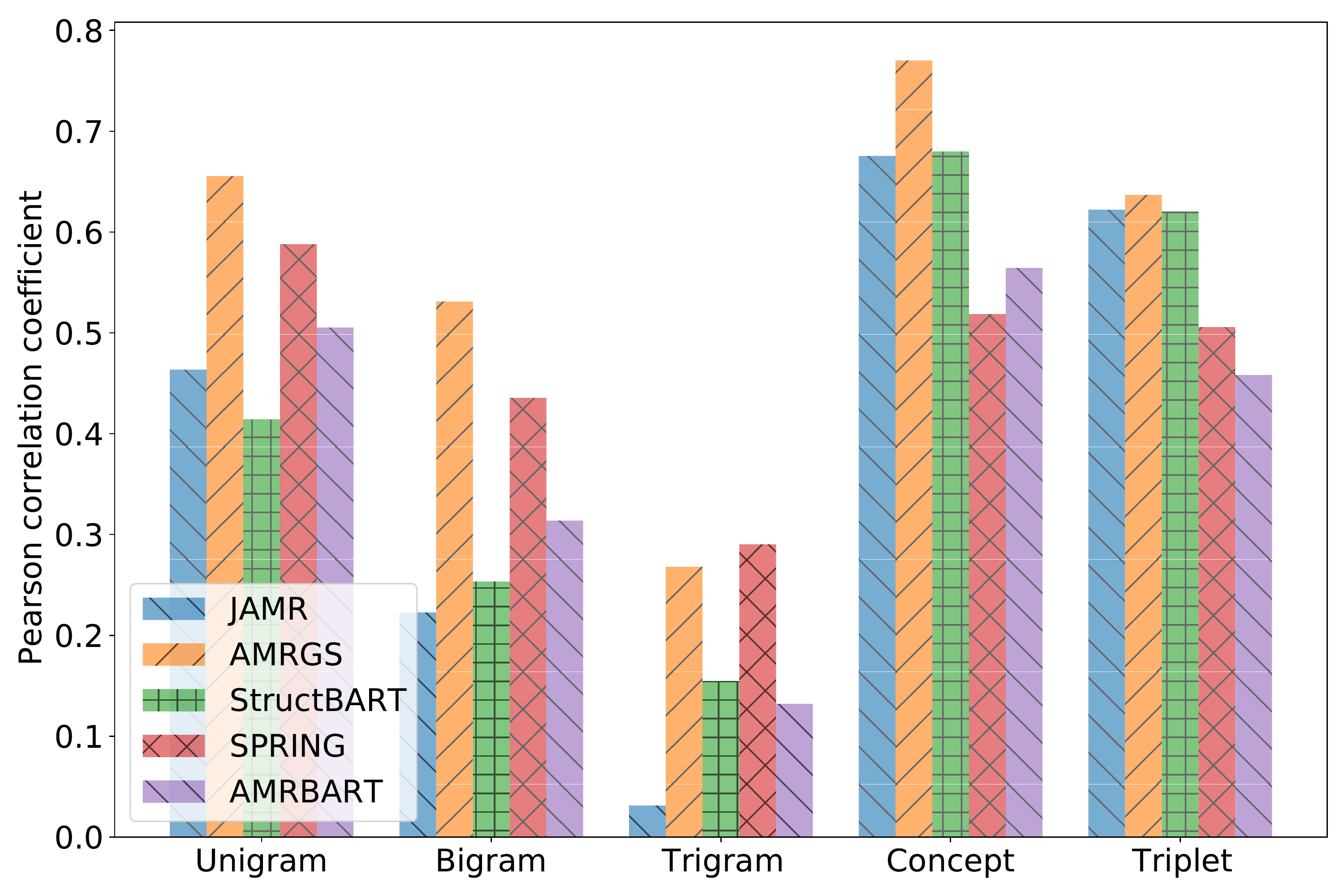}\label{fig:corr-oov}}
	\vspace{-10pt}
	\subfigure[]{\includegraphics[width=0.9\hsize]{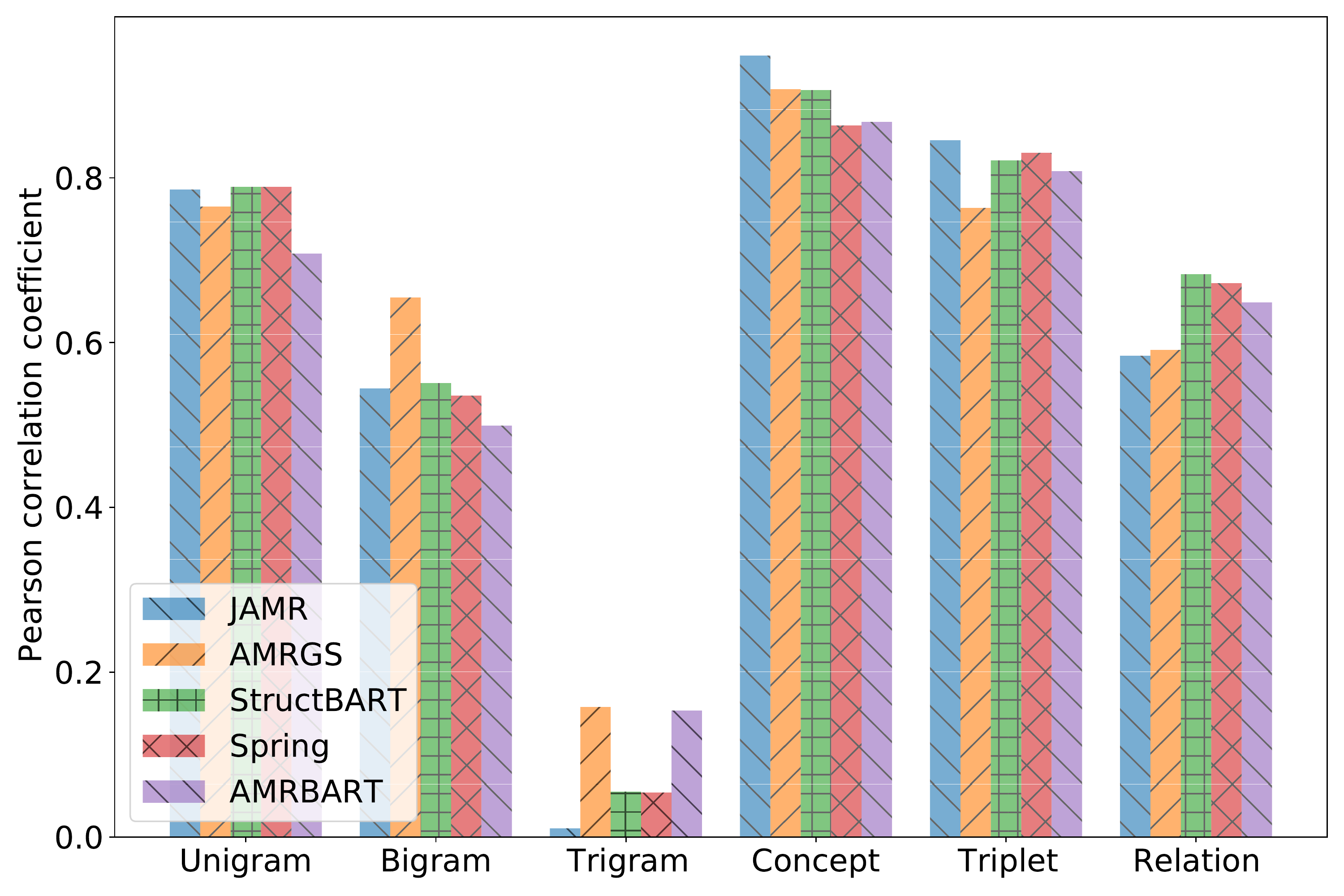}\label{fig:corr-dist}}
    \caption{Pearson correlation coefficient between performance (\textsc{Smatch}) degradation rate and difference of feature distribution measured by (a) OOV rate; (b) Jensen-Shannon divergence. We do not include relation in sub-figure (a) because the relation OOV rate of most test sets are zeros.}
    \label{fig:corr}
    % \vspace{-10pt}
\end{figure}

\subsection{Feature Analysis}       % 从特征层面，待修改
\label{sec:factors}
To study the key factors that impact cross-domain AMR parsing performance, we measure the Pearson correlation coefficient between a set of linguistic features (as introduced in Section~\ref{sec:Discrepancy}) and the relative performance degradation rate.
To eliminate the influences of domain-specific features\footnote{For example, QALD-9 has the smallest averaged input length among all domains, so the  relatively high OOD performance on QALD-9 does not imply that its concept / relation gives smaller domain shift than other domains do. }, we concatenate all OOD test sets and apply bootstrapping procedure~\cite{efron1994introduction,koehn-2004-statistical} to obtain a number of simulated test sets, which are taken as samples for calculating the correlation scores.
Specifically, we create 100 homologous test sets, each with 2,000 examples (out of 3,147) sampled from the concatenated set.
We consider both Jensen-Shannon divergence and OOV rate to measure feature differences.
In Figure~\ref{fig:corr}, each group of columns shows the linear correlation coefficient between the domain divergences of a specific feature (i.e., Table~\ref{tab:dataset-diff}) and the cross-domain performance degradation rates of a specific parser (i.e., Table~\ref{tab:main-parsing}).
We have the following observations:

\begin{itemize}
    \item 
    It can be observed that all parsers are more influenced by domain shift of uni-gram token features while less by those of more complex token features such as bi-gram and tri-gram.
    % This observation is similar to the findings on syntactic parsing~\cite{yang-etal-2022-challenges}.
    The reason might be that AMR parsers rely more on the particular token itself rather than its context for concept identification.
    \item 
    Concepts have larger influences on parsers' performances than relations, indicating that concept identification is the main bottleneck for cross-domain AMR parsing.
    % {This also implies that relation differences are relatively smaller than concept differences, which is consistent with our observation in Table~\ref{tab:dataset-diff}.}
    \item 
    Compared with input textual features, all parsers are more influenced by output AMR structural features, which is consistent with findings of \citet{yang-etal-2022-challenges} and \citet{cui-etal-2022-investigating} in constituency parsing. .
    % \sen{This may explain why \textsc{StructBART} shows lower in-domain performance than SPRING does but achieves higher OOD generalization results (i.e., lower relative performance reduction rates). \textsc{StructBART} explicitly models AMR graph structures thourgh left- and right-arc actions and thus is more suitable for OOD generalization. }
    % \item 
    % Apart from relation, the cross-domain performance is more correlated with Jensen-Shannon divergence compared with OOV rate.
    % This shows that xx.
    %\bai{This suggest that models which pay more attention to AMR structures are more suitable for cross-domain AMR parsing.}
    This suggest that future cross-domain AMR parsing systems should pay more attention on AMR structures.
\end{itemize}

\section{Bridging the Domain Gap}
According to our analysis in Section~\ref{sec:factors}, we investigate two approaches to improve the model performance on OOD datasets by bridging the distribution gap between the training and test domains without modifying model structures.

\noindent \textbf{Alleviating Input Feature Divergence}.
% \noindent \textbf{Domain Adaptive Pre-training}.
% We employ raw text from target domain to reduce the distribution divergence of input features.
We employ raw text from target domain to enrich the model with domain-specific input features.
Specifically, we collect raw text from the biomedical and fairy tales domain, which are then used as extra knowledge for training.
Inspired by previous work~\cite{gururangan-etal-2020-dont}, we add an intermediate pre-training step to adapt the pre-trained model to the target domain, which refers to domain-adaptive pre-training.
We take BART~\cite{lewis-etal-2020-bart} as the backbone and continually pre-train BART on the collected dataset using the standard self-supervised learning training objective.
We randomly mask text spans, replacing $15\%$ tokens.
The adaptively trained model is used for initialization during fine-tuning.

\begin{table}
	\centering
	\small
	\begin{tabular}{lcc}
		\toprule
        \textbf{Model} & \textbf{Bio} & \textbf{TLP} \\
        % \midrule
        % \textit{Original uni. (con.) diver.} &0.28 (0.44) &0.18 (0.30) \\
		\midrule
		\multicolumn{3}{l}{\textit{\textbf{With OOD raw data}}}  \\
		\textit{Unigram/Concept diver.} &0.28/0.44 &0.18/0.30 \\
		\textsc{StructBART} & \textbf{61.2} (+0.8) & \textbf{80.7} (+0.5) \\
		\textsc{SPRING} & 61.0 (+1.3) & 80.4 (+0.5) \\
% 		\textsc{AMRBART} & 63.4 (+0.2) & 82.0 (+0.4) \\
		\midrule
		\textit{\textbf{With OOD silver data}} & &\\
		\textit{Unigram (Concept) diver.} &0.28 (0.30) & 0.18 (0.22) \\
		\textsc{StructBART} & 62.8 (+2.4) &\textbf{81.3} (+1.1) \\
		\textsc{SPRING} &\textbf{63.0} (+3.3) &81.1 (+1.2) \\
% 		\textsc{AMRBART} &63.4 (+0.2) & 82.1 (+0.5) \\
		\bottomrule
	\end{tabular}
	\caption{\textsc{Smatch} score (and improvements) on Bio and TLP when training with out-of-domain data. ``diver.''-- Jensen-Shannon divergence.}
	\label{tab:dapt}
% 	\vspace{-10pt}
\end{table}

\noindent \textbf{Alleviating Output Feature Divergence}.
%\noindent \textbf{Fine-tuning with Cross-domain Silver Data.} 
We investigate silver data as pseudo target domain training data to fine-tune the AMR parsers.
In this way, we expect the cross-domain distribution divergence of both text and AMR features can be reduced. 
We construct the silver data by applying a pre-trained AMR parser to parse collected domain-specific data into AMR graphs.
We use a mixture of gold and silver data to train the models.

\noindent \textbf{Results}. Table~\ref{tab:dapt} shows the results of two BART-based systems\footnote{We do not consider AMRBART, because AMRBART has been trained using large-scale silver data.} on Bio and TLP.
First, with domain-specific raw data, the Jensen-Shannon divergence of unigram reduces significantly ($p$\textless0.01) compared with Table~\ref{tab:dataset-diff}, reaching 0.28 and 0.18 on Bio and TLP, respectively.
Both parsers give better results when initialized with the adaptively pre-trained model.
This confirms our assumption that reducing the input distribution gap can benefit cross-domain AMR parsing.
In addition, the distribution divergence of both unigram and concept decrease when using domain-specific silver data, and both models obtain significant improvements, with a large margin of $2.4$ and $3.3$ points on Bio.
This suggests that reducing distribution divergence of AMR features can also lead to better results.
Finally, compared with input textual features, AMR features give larger improvements.
This is consistent with our observations in Section~\ref{sec:factors}.

\begin{figure}[!t]
    \centering
    \includegraphics[width=0.95\hsize]{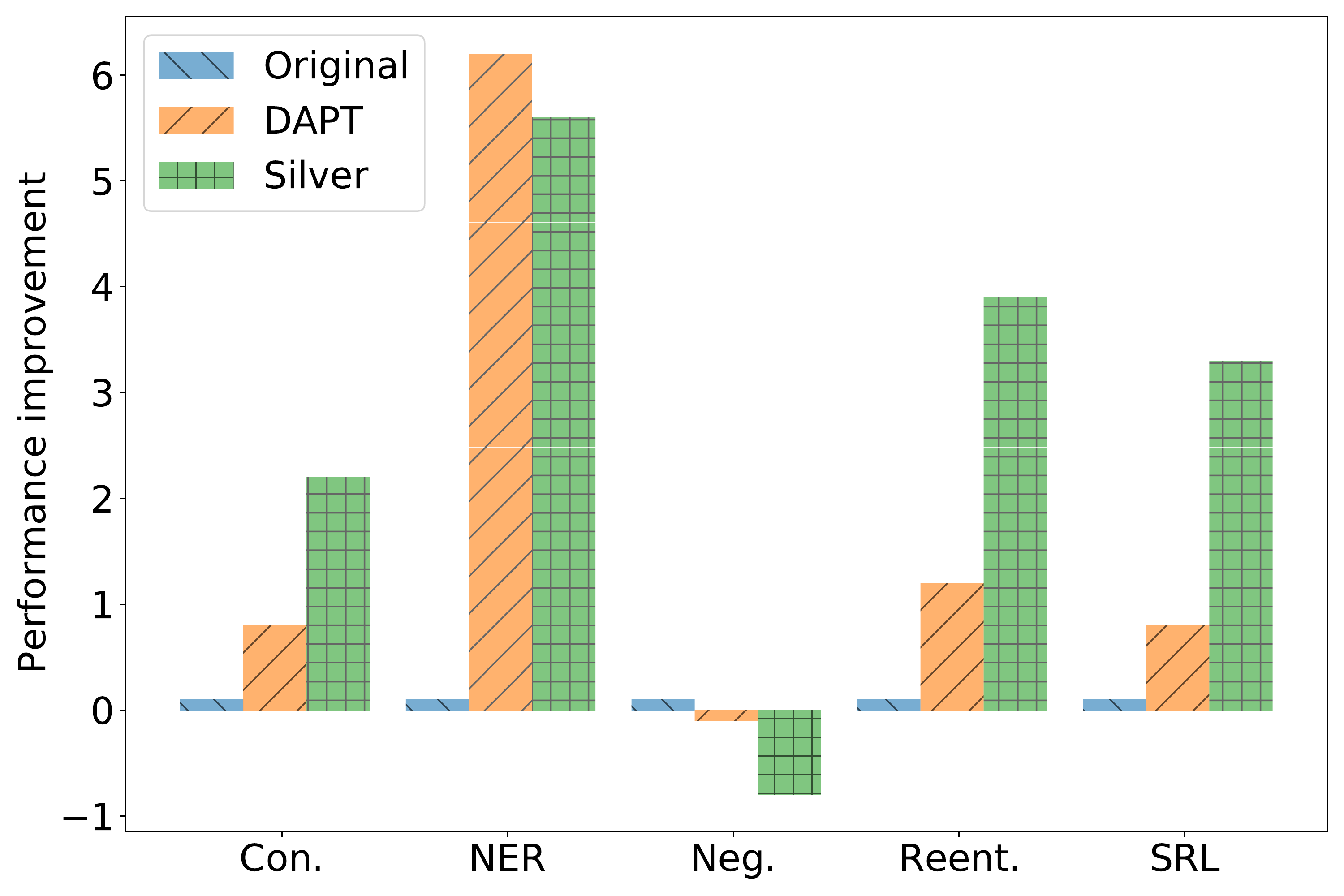}
    \caption{Performance improvements regarding fine-grained evaluation metrics. ``Con.''--Concept, ``Neg.''--Negation, ``Reent.''--Reentrancy.}
    \label{fig:ablation-by-metrics}
    % \vspace{-10pt}
\end{figure}

\noindent\textbf{Fine-grained Evaluation}. Figure~\ref{fig:ablation-by-metrics} shows the fine-grained evaluation results on Bio. 
We take the original SPRING model (Original) as a baseline and compare the performance with the model augmented by domain adaptive pre-training (DAPT) and silver data (Silver).
It can be observed that both methods improve the NER score over the baseline by a large margin (up to $6$ points). 
Also, both methods give better results on concept identification, reentrancy detection, and semantic role labeling.
Compared with DAPT, Silver obtains significantly better results on graph-aware metrics (i.e., concept, reentrancy and SRL), showing that silver data can improve the model performance on predicting structures.
The results of Silver is weaker than DAPT on text-related metrics (i.e., NER and negation).
A possible reason is that silver data might contain noise, which hinders the model to make predictions from textual features.

\section{Conclusion}
We investigated the cross-domain generalization challenges for AMR parsing by analyzing the performance of five representative models. 
Empirically, we found that all AMR parsers give lower performance on OOD test sets, and
the difficulty lies more in output features divergences, including concept and relation, compared with input features. 
% Based on our analysis, we investigated a continual pre-training method, which utilizes the plain text and silver AMR graph on target domain, to bridge the domain gap of input features and output features, respectively.
Based on our analysis, we investigated two approaches to bridge the domain gap of input and output features, respectively, which achieve higher scores on out-of-domain test sets than previous work.
In the future, we would like to investigate more methods, such as vocabulary adaptation~\cite{sato-etal-2020-vocabulary} and k-nearest-neighbor (KNN, ~\citealt{Khandelwal2020Generalization,khandelwal2021nearest}), to improve cross-domain AMR parsing.

\section*{Limitations}
The limitation of our work can be stated from three perspectives. First, the proposed methods do not improve the in-domain parsing performance. Second, we only analyze the cross-domain performance of five representative AMR parsers. Third, we focus on cross-domain AMR parsing in one major language. The performance of other languages remains unknown.

\section*{Acknowledgments}
% \textcolor{blue}{
Yue Zhang is the corresponding author. 
We would like to thank anonymous reviewers for their insightful comments and Yuchen Niu for his help in preliminary experiments.
This work is supported by the National Natural Science Foundation of China
under grant No.61976180 and the Tencent AI Lab Rhino-Bird Focused Research Program.

\bibliography{output}
\bibliographystyle{acl_natbib}
\clearpage
\appendix

\section{Appendix}
\label{sec:appendix}
\subsection{Fine-grained Evaluation Metric for AMR Parsing}\label{appendix:metric}
The Smatch score~\cite{cai-knight-2013-smatch} measures the degree of overlap between the gold and the prediction AMR graphs. 
It can be further broken into different sub-metrics, including:
\begin{itemize}
    \item Unlabeled (Unlab.): Smatch score after removing edge-labels
    \item NoWSD: Smatch score after ignoring Propbank senses (\emph{e.g.}, go-01 vs go-02)
    \item Concepts (Con.): $F$-score on the concept identification task
    \item Wikification (Wiki.): $F$-score on the wikification (\texttt{:wiki} roles)
    \item Named Entity Recognition (NER): $F$-score on the named entities (\texttt{:name} roles).
    \item Reentrancy (Reen.): Smatch score on reentrant edges.
    \item Negation (Neg.): $F$-score on the negation detection (\texttt{:polarity} roles).
    \item Semantic Role Labeling (SRL): Smatch score computed on~\texttt{:ARG-i} roles.
\end{itemize}

\begin{table*}[!b]
	\centering
	\small
	\begin{tabular}{l|c|cccrccrc}
		\toprule
        \textbf{Model} & \textbf{Smatch} & \textbf{Unlab.}  & \textbf{NoWSD} & \textbf{Con.} &\textbf{Wiki.} & \textbf{NER} & \textbf{Reent.} & \textbf{Neg.} & \textbf{SRL}\\
		\midrule
		\textbf{AMR2.0 (ID)} & & & & & & & & &\\
		JAMR &67.0 &71.6 &67.7 &83.0 &75.9 &80.3 &61.0 &43.9 &59.7 \\
		AMRGS & 80.6	&83.9 &	81.0	&88.1	&86.5	&81.1	&64.7	&78.5	&74.3\\
		StructBART &84.1 &87.6 &84.4 &90.4 &79.6 &\textbf{92.2} &\textbf{74.3} &71.2 &\textbf{83.0} \\
		SPRING &84.7    &87.6 &84.9     &90.2 &\textbf{87.3}        &83.7   &72.3   &\textbf{79.9}  &79.7 \\
		AMRBART &\textbf{85.5} &\textbf{88.4} &\textbf{85.9} &\textbf{91.2} & 84.4 &{91.5} &{73.5} &73.5 &{81.5} \\
	   \midrule
		\textbf{New3 (OOD)} & & & & & & & & &\\
		JAMR &57.2	&62.5	&57.8	&73.1	&49.8	&52.7	&38.9	&28.3	&53.2 \\
		AMRGS &61.8	&66.8	&62.2	&75.9	&49.6	&45.4	&54.8	&59.6	&65.0 \\
		StructBART &74.0	&78.1 &74.5	&83.1	&53.6	&71.1	&63.2	&63.3	&72.1 \\
		SPRING &74.2	&78.4	&74.6	&82.3	&60.1	&66.4	&62.9	&64.2	&71.7 \\
		AMRBART &\textbf{77.3}	&\textbf{81.2}	&\textbf{77.8}	&\textbf{84.6}	&\textbf{73.5}	&\textbf{72.0}	&\textbf{65.6}	&\textbf{66.7}	&\textbf{73.7} \\
		\midrule
		\textbf{TLP~\textit{v1.6} (OOD)} & & & & & & & & &\\
		JAMR &59.9	&66.7	&60.9	&88	 &25.5	&53.0   &32.4	&55.4	&54.6 \\
		AMRGS &73.7	&78.4  &74.6	&82.4	&33.1	&24.1	&58.7	&63.5	&70.8 \\
		StructBART &80.2	&84.3	&81.0	 &87.1	&69.5	&\textbf{75.2}	&67.9	&77.3	&77.4 \\
		SPRING &79.9 &83.9	&80.7  &86.4	&65.7	&63.2	&67.0	&\textbf{80.9}	&77.0 \\
		AMRBART &\textbf{81.6}	&\textbf{85.3}	&\textbf{82.3}	&\textbf{87.8}	&\textbf{87.4}	&73.5	&\textbf{69.3}	&77.8	&\textbf{78.7} \\
		\midrule
		\textbf{TLP~\textit{v3.0} (OOD)} & & & & & & & & &\\
		JAMR &58.8	&66.0	&59.7	&75.9	&25.5	&53.0   &31.7	&49.1	&52.9 \\
		AMRGS &72.0	&77.0	&72.9	&81.3	&33.1	&24.1	&57.2	&57.5	&68.5 \\
		StructBART &78.5	&83.0	&79.2	&85.9	&69.5	&\textbf{75.2}	&66.1	&70.1	&75.0 \\
		SPRING &78.2 &82.6	&79.0 	&85.3	&65.7	&63.2	&65.0	&\textbf{72.9}	&74.7 \\
		AMRBART &\textbf{79.8}	&\textbf{84.0}	&\textbf{80.5}	&\textbf{86.7}	&\textbf{84.7}	&73.5	&\textbf{67.6}	&70.7	&\textbf{76.4} \\
		\midrule
		\textbf{Bio~\textit{v0.8} (OOD)} & & & & & & & & &\\
		JAMR &38.7	&44.1 &39.6	 &56.9	&~7.6	&15.6	&26.0	&50.3	&37.3 \\
		AMRGS &43.9	&49.8	&44.4	&55.6	&~\textbf{9.0}	&~6.4	&34.1	&60.4	&47.0 \\
		StructBART &60.4	&64.9	&60.9	&70.1	&~1.9	&31.9	&43.6	&76.0	&56.7 \\
		SPRING &59.7	&63.7	&60.2	&71.1	&~3.2	&33.7	&43.5	&\textbf{75.7}	&57.5 \\
		AMRBART &\textbf{63.2}	&\textbf{67.2}	&\textbf{63.9}	&\textbf{73.4}	&~2.0   &\textbf{39.7}	&\textbf{47.1}	&75.4	&\textbf{60.6} \\
		\midrule
		\textbf{Bio~\textit{v3.0} (OOD)} & & & & & & & & &\\
		JAMR &38.4	&43.9 &39.3	 &56.7	&~7.5	&15.6	&25.8	&44.9	&36.7 \\
		AMRGS &43.2	&49.1	&43.7	&55.0	&~\textbf{8.6}	&~6.4	&33.8	&55.6	&45.9 \\
		StructBART &57.6	&62.0	&58.2	&69.2	&~1.9	&31.9	&42.9	&70.1	&55.2 \\
		SPRING &59.2	&63.5	&59.6	&70.2	&~3.2	&33.7	&43.2	&\textbf{72.4}	&56.2 \\
		AMRBART &\textbf{62.1}	&\textbf{66.3}	&\textbf{62.9}	&\textbf{72.7}	&~2.0   &\textbf{39.7}	&\textbf{46.7}	&70.5	&\textbf{59.2} \\
		\midrule
		\textbf{QALD-9 (OOD)} & & & & & & & & &\\
		JAMR &60.8	&66.6	&61.3	&69.9	&~~0	&61.2	&34.2	&5.0	&56.4 \\
		AMRGS &70.0	&74.5	&70.2	&80.1	&~~0	&48.7	&57.0	&~9.5	&72.0 \\
		StructBART &83.7    &86.9	&83.9	&89.7	&~~0	&81.9	&61.3	&14.0   &76.3 \\
		SPRING &80.4	&83.2	&80.6	&88.9	&~~0	&80.4	&56.5	&~9.3	&75.4 \\
		AMRBART &\textbf{85.1}	&\textbf{87.3}	&\textbf{85.2}	&\textbf{91.8}	&~~0	&\textbf{83.8}	&\textbf{71.9}	&\textbf{69.1}	&\textbf{83.6} \\
		\bottomrule
	\end{tabular}
	\caption{AMR parsing results on in-domain and out-of-domain test sets. The best results within each row block are shown in bold.}
	\label{tab:main-parsing-appendix}
% 	\vspace{-0.7em}
\end{table*}
\subsection{Full Cross-domain Performance}
\label{sec:appendix-more-res}
Table~\ref{tab:main-parsing-appendix} shows the detailed results of AMR parsing on different test sets in terms of $9$ evaluation metrics.
\end{document}